\documentclass[11pt]{article}
\usepackage{isolatin1}
\usepackage{acl04}
\usepackage{balance} 
\usepackage{eso-pic} 
\usepackage{graphicx} 
\AddToShipoutPicture{\put(50,750){\texttt{Second International Workshop on Evaluating Word Sense Disambiguation Systems (SENSEVAL) 2001}}}

\title{The UNED systems at \sc{Senseval-2}}

\author{David Fernández-Amorós, Julio Gonzalo, Felisa Verdejo \\
  Depto. de Lenguajes y Sistemas Informáticos, UNED \\
  \{david,julio,felisa\}@lsi.uned.es} 

\begin{document}
\maketitle
\begin{abstract}

We have participated in the \textsc{Senseval-2} English tasks (all words and
lexical sample) with an unsupervised system based 
on mutual information measured over a large corpus (277 million
words) and some additional heuristics. A supervised extension of the
system was also presented to the lexical sample task. 

Our system scored first among unsupervised systems in both tasks: 56.9\%
recall in all words, 40.2\% in lexical sample. This is slightly worse
than the 
first sense heuristic for all words and 3.6\% better for the lexical sample, a
strong indication that unsupervised Word Sense Disambiguation remains
being a strong challenge.

\end{abstract}

\section{Introduction}

  We advocate researching unsupervised techniques for
  Word Sense Disambiguation (WSD). Supervised techniques offer better results in general but the
  setbacks, such as the problem of developing reliable training data,
  are very considerable. Also there's probably more to WSD than blind
  machine learning (a typical approach, although such systems produce
  interesting
  baselines). 

Within the unsupervised paradigm, we are interested in
performing in-depth measures of the disambiguation potential of
different sources of information. We have previously investigated the
informational value of semantic distance measures in
\cite{Fernandez-Amoros-01a}. For \textsc{Senseval-2}, we have turned to
investigate pure coocurrence information as a source of disambiguation 
evidence.
In essence, our system computes a matrix of mutual
  information for a fixed vocabulary and applies it to weight
  coocurrence counting between sense and context characteristic
  vectors. 

In the next section we describe the process of constructing
  the relevance matrix. In section 3 we present the particular
  heuristics used for the competing systems. In section 4 we show the
  results by system and heuristic and some baselines for
  comparison. Finally in the last sections we draw some conclusions
  about the exercise.

\section{The Relevance matrix}
\subsection{Corpus processing}

  Before building our systems we have developed a resource we've
  called the \emph{relevance matrix}. The raw data used to build the matrix
  comes from the Project Gutenberg
  (PG)~\footnote{\texttt{http://promo.net/pg}}.  

  At the time of the creation of
  the matrix the PG consisted of more than 3000 books of diverse
  genres. We have 
  adapted these books for our purpose~: First, language identification
  was used to filter books written in English; Then we stripped
  off the disclaimers. The result is a collection of around 1.3Gb of
  plain text.

  Finally we tokenize, lemmatize, strip
  punctuation and stop words and detect numbers and
  proper nouns.

\subsection{Coocurrence matrix}

  We have built a vocabulary of the 20000 most frequent words (or
  labels, as we have changed all the proper nouns detected to the label
  PROPER\_NOUN and all numbers detected to NUMBER) in the
  text and a symmetric coocurrence matrix between these words within a
  context of 61 words (we thought a broad context of radius 30 would
  be appropriate since we are trying to capture vague semantic relations).

\subsection{Relevance matrix}

  In a second step, we have built another symmetric matrix, which we have
  called \emph{relevance matrix}, using a mutual information measure
  between
  the words (or labels), so that for two words $a$ and $b$, the entry
  for them would be $ \frac{P(a \cap b)}{P(b)P(a)}$, where $P(a)$ is
  the probability of finding the word $a$ in a random context of a
  given size. $P(a \cap b)$ is the probability of finding both $a$
  and $b$ in a random context of the fixed size. We've introduced a
  threshold of 2 below which we set the entry to zero for practical
  purposes. We think that this is a valuable resource that could
  be of interest for many other applications other than WSD. Also, it
  can only grow in quality since at the time of making this report the data
  in the PG has almost doubled in size.

\section{Cascade of heuristics}

 We have developed a very simple language in order to systematize the
 experiments. This language allows the construction of WSD systems
 composed of different heuristics that are applied in cascade so that
 each word to be disambiguated is presented to the first heuristic,
 and if it fails to disambiguate, then the word is passed on to the 
 second heuristic and so on. We can have several such systems running in
 parallel for efficiency reasons (the  matrix has high memory
 requirements). Next we show the heuristics we have considered to build
 the systems

\begin{itemize}
\item \textbf{Monosemous expressions}.

  Monosemous expressions are simply unambiguous words in the case of
  the all words English task.
  In the case of the lexical sample English task, however, the
  annotations include multiword expressions. We have
  implemented a multiword term detector that considers the multiword
  terms from WordNet's index.sense file and 
  detects them in the test file using a multilevel backtracking
  algorithm that takes account of the inflected and base forms of the
  components of a particular multiword in order to maximize
  multiword detection. We tested this algorithm against the
  PG and found millions of these multiword terms.

  We restricted ourselves to the multiwords already present in the
  training file since there are, apparently, multiword expressions
  that where overlooked during manual tagging (for instance the WordNet
  expression 'the\_good\_old\_days' is not hand-tagged as such in the
  test files)

\item \textbf{Statistical filter}

  WordNet comes with a file, cntlist, literally 'file listing number
  of times each tagged sense occurs in a semantic concordance' so we
  use this to compute the relative probability of a sense
  given a word (approximate in the case of collections other than
  SemCor). Using this information, we eliminated
  the senses that had a probability under 10\% and if only one sense
  remains we choose it. Otherwise we go on to the next
  heuristic. In other words, we didn't apply complex techniques
  with words which are highly skewed in meaning~\footnote{Some people
    may argue that this is a supervised approach. In our opinion, the
    cntlist information does not make a system supervised per se,
    because a) 
    It is standard information provided as part of the dictionary and
    b) We don't use the examples to feed or train any procedure.}.

\item \textbf{Relevance filter}

  This heuristic makes use of the relevance matrix. In order to assign
  a score to a sense, we count the coocurrences of words in the context
  of the word to be disambiguated with the words in the definition
  of the senses (the WordNet gloss tokenized, lemmatized and stripped
  out of stop words and 
  punctuation signs) weighting each coocurrence by the entry in the 
  relevance matrix for the word to be disambiguated and the word whose
  coocurrences are being counted, i.e., if $s$ is a sense of the
  word $\alpha$ whose definition is $S$ and $C$ is the
  context in which 
  $\alpha$ is to be disambiguated, then the score for $s$ would be:

$$\sum_{w \in C}{R_{w \alpha} \mbox{freq}(w,C) 
  \mbox{freq}(w,S) \mbox{idf}(w,\alpha)}$$

  Where idf$(w,\alpha) = \log \frac{N}{d_w}$, with $N$ being the
  number of senses for word $\alpha$ and $d_w$ the number of sense glosses in
  which $w$ appears. $\mbox{freq}(w,C)$ is the frequency of word $w$ in the
  context $C$ and $\mbox{freq}(w,S)$ is the frequency of $w$ in the sense
  gloss $S$. 

  The idea is to prime the occurrences of words that are relevant to
  the word being disambiguated and give low credit (possibly none) to
  the words that are incidentally used in the context.

  Also, in the all words task (where POS tags from the TreeBank are
  provided) 
  we have considered only the context words that have a
  POS tag compatible with that of the word being disambiguated. By
  compatible we mean nouns and nouns, nouns and verbs, nouns and
  adjectives,  verbs and verbs, verbs and adverbs and vice
  versa. Roughly speaking, words that can have an intra-phrase
  relation.

  We also filtered out senses with low values in the cntlist file,
  and in any case we only considered at most the first six senses of a
  word.

\item \textbf{Enriching sense characteristic vectors}

  The relevance filter provided very good results in  our experiments
  with SemCor and \textsc{Senseval-1} data as far as precision is concerned, but
  the problem is that there is little overlapping between the
  definitions of the senses and the contexts in terms of coocurrence
  (after removing stop words and computing idf) which means that the
  previous heuristic didn't disambiguate many words. 

  To overcome this problem, we enrich the senses characteristic
  vectors adding for each word in the vector the words related to it
  via the relevance matrix weights. This corresponds to the
  algebraic notion of multiplying the matrix and the characteristic
  vector. In other words, if $R$ is the relevance matrix and $v$ our
  characteristic vector we would finally use $Rv + v$.

  This should increase the number of words disambiguated
  provided we eliminate the idf factor (which would be zero
  in most cases because now the sense characteristics vectors are not as
  sparse as before). When we also discard senses with low relative
  frequency in SemCor we call this heuristic \emph{mixed filter}.

\item \textbf{back off strategies}

  For those cases that couldn't be covered by other heuristics we
  employed the first sense heuristic. In the case of the supervised
  system for the English lexical sample task we thought of using the
  most frequent sense but didn't implement it due to lack of time.

\end{itemize}

\section{Systems and Results}

\begin{itemize}
\item \textbf{UNED--AW--U2}

  We won't delve into UNED-AW-U system as it is very similar to this
  one. This is an (arguably) unsupervised system for the English all
  words task. 
  The heuristics we used and the results obtained for each of them are
  shown in Table 1.

\begin{table}[htbp]
{\small
  \begin{center}
    \begin{tabular}{||l|c|c|c|c|} \hline
      Heuristic              & Att. & Score & Prec & Rec\\ \hline
      Monosemous exp & 514       & 45500 & 88.5\%    & 18.4\%\\ \hline
      Statistical filter     & 350       & 27200 & 77.7\%    & 11.0\%\\ \hline  
      Mixed filter & 1256 & 50000 & 39.8\% & 20.2\%\\ \hline
      Enriched Senses        & 77        & 4300  & 55.8\%    & 3.1\% \\ \hline
      First sense            & 249       & 13600 & 54.6\%    & 5.5\%\\ \hline
      Total                  & 2446      & 140600 & 57.5\%   & 56.9\%\\ \hline
    \end{tabular}
    \caption{Unsupervised heuristics for English all words task}
   \end{center}
}
\end{table}

 If the individual heuristics are used as standalone WSD systems we
 would obtain the results in Table 2.

\begin{table}[htbp]
{\small
  \begin{center}
    \begin{tabular}{||l|c|c|c|c|} \hline
      System                   & Att.  & Score & Prec & Recall \\ \hline
      First sense & 2405       & 146900  & 61.1\%  & 59.4\%      \\ \hline
      UNED--AW--U2 & 2446      & 140600  & 57.5\%  & 56.9\%      \\ \hline
      Mixed filter     & 2120    & 122600  & 57.8\% & 49.6\%\\
      \hline
      Enriched senses          & 2122    & 108100    & 50.9\%    & 43.7\% \\ \hline
      Random       & 2417      & 89191.2 & 36.9\%  & 36.0\%      \\ \hline
      Statistical filter       & 864     & 72700   & 84.1\%    & 29.4\%\\ \hline  
    \end{tabular}
    \caption{UNED--AW--U2 vs baselines}
   \end{center}
}
\end{table}

\end{itemize}

 In the lexical sample task, we weren't able to multiply by the
 relevance matrix due to time constraints, so in order
 to increase the coverage for the relevance filter heuristic we
 expanded the definitions of the senses with those of the first 5 levels
 of
 hyponyms. Also, we selected the radius of the context to be
 considered depending on the POS of the word being disambiguated. 
 For nouns and verbs we used 25 words radius neighbourhood and for
 adjectives 5 words at each side.  

\begin{itemize}

\item \textbf{UNED--LS--U}
   This is essentially the same system as UNED--AW--U2, in this case
   applied to the lexical sample task. The results are displayed in
   Table~3.

\begin{table}[htbp]
{\small
  \begin{center}
    \begin{tabular}{||l|c|c|c|c|} \hline
      Heuristic                 & Att.      & Score  & Prec      & Recall    \\ \hline
      Relevance filt          & 3039      & 113617 & 37.3\%    & 26.2\%    \\ \hline
      First sense               & 1285      & 60000  & 46.7\%    & 13.9\%    \\ \hline
      Total                     & 4324      & 173617 & 40.2\%    & 40.2\%    \\ \hline
    \end{tabular}
    \caption{Unsupervised heuristics for English lexical sample task}
   \end{center}
}
\end{table}

\item \textbf{UNED--LS--T}

 This is a supervised variant of the previous systems. We have added
 the training examples to the definitions 
 of the senses giving the same weight to the definition and to all the
 examples as a whole (i.e. definitions are considered more interesting
 than examples)

\begin{table}[htbp]
{\small
  \begin{center}
    \begin{tabular}{||l|c|c|c|c|} \hline
      Heuristic                 & Att.      & Score  & Prec      & Recall    \\ \hline
      Relevance filt          & 4116      & 206150 & 50.1\%    & 47.6\%    \\ \hline
      First sense               & 208       & 9300   & 44.7\%    & 2.1\%     \\ \hline
      Total                     & 4324      & 215450 & 49.8\%    & 49.8\%    \\ \hline
    \end{tabular}
    \caption{Supervised heuristics for English lexical sample task}
   \end{center}
}
\end{table}

\end{itemize}

\section{Discussion and conclusions}

  We've put a lot of effort into making the relevance matrix but its
  performance in the WSD task is striking. The matrix is interesting
  and its application in the relevance filter heuristic is slightly
  better than simple coocurrence counting, which proves that it
  doesn't discard relevant words. The problem seems to lie in the fact
  that irrelevant words (with respect to the word to be disambiguated) rarely
  occur both in the context of the word and in the definition of the
  senses (if they appeared in the definition they wouldn't be so irrelevant)
  so the direct impact of the information in the matrix is very
  weak. Likewise, relevant (via the matrix) words with respect to the word to
  be disambiguated occur often both in the context and in the
  definitions so the final result is very similar to simple
  coocurrence counting. 

  This problem only showed up in the lexical sample task systems. In
  the all words systems we were to enrich the sense definitions to
  make a more advantageous use of the matrix.

  We were very confident that the relevance filter would yield good
  results as we have already evaluated it against the \textsc{Senseval-1} and
  SemCor data. We felt however that we could improve the coverage of
  the heuristic enriching the definitions multiplying by the matrix. A
  similar approach was used by Yarowsky \cite{Yarowsky-92} and Schütze
  \cite{Schuetze-95} and it 
  worked for them. This wasn't the case for us; still, we think the
  resource is well worth researching other ways of using it. 

  As for the overall scores, the unsupervised lexical sample obtained
  the highest recall of the unsupervised  systems, which
  proves that carefully 
  implementing simple techniques still pays off. In the all words task
  the UNED-WS-U2 had also the highest recall among the
  unsupervised systems (as characterized in the \textsc{Senseval-2}
  web descriptions), and the fourth overall. We'll train it with
  the examples in Semcor 1.6 and see how much we can gain. 

\section{Conclusions}

Our system scored first among unsupervised systems in both tasks: 56.9\%
recall in all words, 40.2\% in lexical sample. This is slightly worse
than the 
first sense heuristic for all words and 3.6\% better for the lexical sample, a
strong indication that unsupervised Word Sense Disambiguation remains
being a strong challenge.

\bibliographystyle{acl}
\bibliography{bibliografia}
\balance 

\end{document}